\documentclass[journal,10pt]{IEEEtran}
\usepackage{amsfonts,graphicx, amsmath, bm,multirow, hhline}
\usepackage{color}
\usepackage{caption}
\usepackage{graphicx}
\usepackage{caption}
\usepackage{amsmath}

\usepackage{subcaption}
\usepackage{subcaption}
\usepackage{graphicx}
\usepackage{amssymb}

\usepackage{times}
\usepackage{verbatim}
\usepackage{cite}
\usepackage{url}
\usepackage{epstopdf}
\usepackage{algorithm}
\usepackage{algorithmicx}
\usepackage{algpseudocode}
\usepackage{amsmath}
\usepackage{bbding}
\newtheorem{theorem}{Theorem}

\newtheorem{remark}{Remark}

\allowdisplaybreaks

\newenvironment{proof}[1][Proof]{\begin{trivlist}
		\item[\hskip \labelsep {\bfseries #1}]}{\end{trivlist}}

\newcommand{\qed}{\nobreak \ifvmode \relax \else
	\ifdim\lastskip<1.5em \hskip-\lastskip
	\hskip1.5em plus0em minus0.5em \fi \nobreak
	\vrule height0.75em width0.5em depth0.25em\fi}

\begin{document}

\title{Energy-Aware Edge Association for Cluster-based Personalized Federated Learning}

\author{Yixuan Li,\mbox{\hspace{0.35cm}} Xiaoqi Qin,\mbox{\hspace{0.35cm}}Hao Chen,\mbox{\hspace{0.35cm}}Kaifeng Han,\mbox{\hspace{0.35cm}} Ping Zhang,~\IEEEmembership{Fellow,~IEEE}

\thanks{Y. Li, X. Qin (\textit{corresponding author}), H. Chen and P. Zhang are with Beijing University of Posts and Telecommunications, (e-mail: lyixuan@bupt.edu.cn; xiaoqiqin@bupt.edu.cn; 2020110229@bupt.edu.cn; pzhang@bupt.edu.cn).}
\thanks{K.Han is with China Academy of Information and Communications Technology (e-mail:hankaifeng@caict.ac.cn).}
}

{}

\maketitle

\begin{abstract}
	
Federated Learning (FL) over wireless network enables data-conscious services by leveraging the ubiquitous intelligence at network edge for privacy-preserving model training.
As the proliferation of context-aware services,
the diversified personal preferences causes disagreeing conditional distributions among user data,
which leads to poor inference performance.
In this sense,
clustered federated learning is proposed to group user devices with similar preference and provide each cluster with a personalized model.
This calls for innovative design in edge association that involves user clustering and also resource management optimization.
We formulate an accuracy-cost trade-off optimization problem by jointly considering model accuracy,
communication resource allocation and energy consumption.
To comply with parameter encryption techniques in FL,
we propose an iterative solution procedure which employs deep reinforcement learning based approach at cloud server for edge association.
The reward function consists of minimized energy consumption at each base station and the averaged model accuracy of all users.
Under our proposed solution,
multiple edge base station are fully exploited to realize cost efficient personalized federated learning without any prior knowledge on model parameters.
Simulation results show that our proposed strategy outperforms existing strategies
in achieving accurate learning at low energy cost.

\end{abstract}


\IEEEpeerreviewmaketitle
\section{Introduction}
Driven by the paradigm shift from ``connected things'' to ``connected intelligence'' towards 6G,
the pattern of social activities is fast forwarded to digital-first mode,
where people are becoming reliant on context-aware support technologies to enjoy personalized services (e.g., smart healthcare, intelligent recommendation) \cite{ZhuGuangxu20:CM:Intelligent-Edge}.
Federated Learning (FL) over wireless network\cite{cui-and-vincent-wirless} is a nascent solution to enable data-conscious and proactive service in a privacy-preserving fashion that the user data does not leave its location \cite{Xiongzehui20:CC:6GFL}.
Note that the essence of FL is to train a high-quality global model which is agreed by all participating users,
using common knowledge extracted from their local data sets \cite{Xujie21:JCIN:Energy-Efficient}.
To combat non-independent identically distributed (non-i.i.d.) data distributions among users,
various techniques have be proposed to enable model fine-tunes at users\cite{Fallah20:arXiv:MFL},
which nevertheless still comes in a generic one-size-fits-all implementations for global model training,
and therefore works well with certain types of non-i.i.d. data such as skewed label distribution.

Considering the fact that users of diversified preferences may locate within the same area served by a base
station.
The differing personal preference results in disagreeing conditional distributions,
and thus one single model would not be able to accurately predict the purpose of all users,
and it is necessary to provide each group of interests with a personalized model that fits their data distribution.
In\cite{Sattler20:TNNLS:Clusteredfederated} and\cite{Briggs20:IJCNN:Clusteredfederated2},
a clustered federated learning framework is proposed to identify the hidden clustering structure among user data,
which improves the model accuracy by minimizing harmful interference between disagreeing users.
Since the clustering is based on distance between model parameters,
it may not be directly applied to secured FL with nonlinear parameter encryption\cite{Mothukuri21:security},\cite{Hao20:PrivacyFL}.

\begin{figure*}[!htbp]
	\centering
	\includegraphics[width=6.8in, clip]{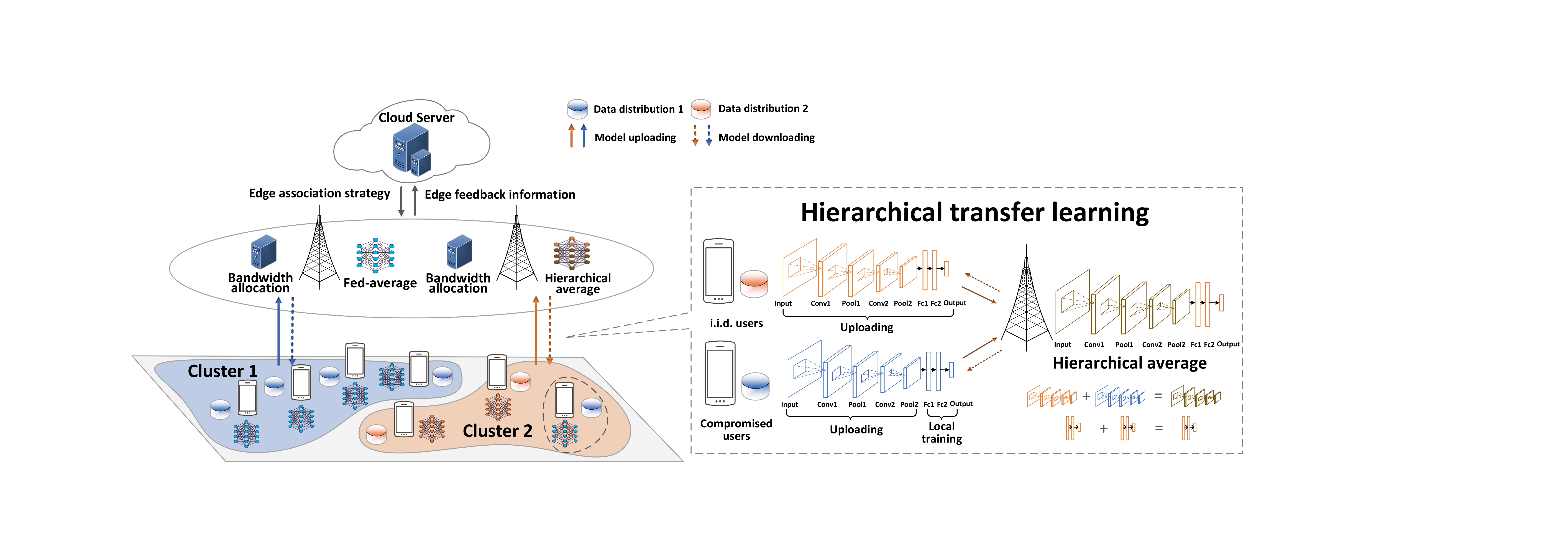}
	\caption{Clustered federated edge learning system.}
	\label{Figure:architecture}
\end{figure*}

Furthermore,
it is essential to take communication cost into consideration when implementing clustered FL over wireless network,
especially for energy-limited users.
User drop-outs due to energy exhaustion during repetitive and collaborative model updates exchanges may be deleterious to training accuracy\cite{Saad-energy}.
Recent works considers model training with single base station where the training performance could be improved by designing data importance-aware user selection strategy\cite{Xiongzehui20:WCom:HFEL} and resource allocation schemes\cite{CUI-PNAS}.
In case when multiple base stations are involved in model training\cite{chenxu20:TWC:HFEL},
dynamic edge association strategies are proposed to find energy efficient matching between users and base stations\cite{Xiongzehui21:JCAC:Dynamic-Association}.
Note that the goal of these works is to train one generic model,
since users are assumed to be congruent (i.e., one central model fits all users' distributions).
Therefore,
personalized model training based on heterogeneity among user data distributions is not considered.

Due to the inconsistency between user data distribution and geographical distribution,
an edge association solution cannot simultaneously serve the purpose of optimizing energy consumption and optimizing model accuracy.
Therefore,
the aforementioned existing solutions cannot be directly applied to find the energy-efficient edge association strategy for clustered federated learning.
It raises a natural question of how to partition the users into multiple clusters for model training,
especially in a privacy-preserving fashion without prior knowledge of user data.
Moreover,
in order to minimize the communication cost of model training with multiple base stations,
it is also important to carefully design the matching between base stations and user clusters,
as well as  bandwidth allocation strategy within each cluster.

In this paper,
we investigate the interplay between model accuracy and communication cost in clustered federated learning over wireless network with multiple base stations.
We formulate a system-level accuracy-energy tradeoff optimization problem by jointly considering the edge association,
bandwidth allocation,
energy consumption,
and model accuracy.
We propose an iterative solution procedure involving edge association at cloud server and resource allocation at base stations.
As for edge association,
we propose an approach based on Deep Reinforcement Learning (DRL) to realize clustering without knowledge on model parameters.
At each base station,
it performs bandwidth allocation among its associated users using convex optimization techniques.
In the case when users with disagreeing distributions are associated with the same base station due to energy constraints,
we adopt hierarchical transfer learning to improve the model accuracy.
Under our proposed scheme,
multiple edge base station are fully exploited  to achieve cost efficient personalized federated learning while preserving user privacy.
\section{Mathematical Modeling and Problem Formulation}
\label{sec:Model}

\subsection{System Architecture}

Consider a clustered federated learning system with multiple edge base stations as shown in Fig.~\ref{Figure:architecture},
which consists of a set of $\mathcal M$ edge base stations, a set of $\mathcal N$ users,
and a cloud server.
Let $M = |\mathcal M|$ denote the number of edge base stations and $N = |\mathcal N|$ denote the number of users.
Assume that users hold arbitrary non-i.i.d. data,
which have disagreeing conditional distributions due to personalized preferences.
Under the guidance of the cloud server,
users with the same conditional distribution will be clustered to associate with an edge base station for personalized model training.
Each edge server allocates communication resource among its associated users to minimize the energy consumption.

\subsection{Clustered Federated Learning Model}
As for edge association,
denote $a_{ij}$ as a binary variable to indicate whether or not user $i$ is associated with base station $j$,
i.e.,
$a_{ij}=1$ if user $i$ is associated with base station $j$ and $0$ otherwise.
Note that each user can be only associated with one base station,
then we have:
\begin{equation}
	\sum\limits_{j} {{a_{ij}} \le 1} ,\hspace{0.3cm} ( {i} \in \mathcal N) \; .
	\label{Eq:aij1}
\end{equation}

At each base station $j$,
it works as an agggregator to train personalized model with its associated users to minimize the loss function:
\begin{equation}
		\mathop {\min }\limits_{{\boldsymbol{\omega} _1},...,{\boldsymbol{\omega} _N}} \left\{F_j\triangleq \sum\limits_{i = 1}^N a_{ij} \frac{1}{|\mathcal D_i|} {\sum\limits_{n = 1}^{{|\mathcal D_i|}}l_if_i({\boldsymbol{\omega} _i},{\boldsymbol{x}_{in}},{y_{in}})}\right\} .
	\label{Eq:flobjective}
\end{equation}
where $f_i({\boldsymbol{\omega} _i},{\boldsymbol{x}_{in}},{y_{in}})$ denotes the local loss function,
and $\mathcal D_i$ is the mini-batch randomly chosen from the data samples of user $i$.

At $t$-th round of training,
user $i$ refines its local model by applying stochastic gradient descent on local data $\mathcal D_i$,
and we have:
\begin{equation}
\begin{split}
{\boldsymbol{\omega} _i}(t + 1) = {{\boldsymbol{\omega}_i}(t)} - \eta \frac{1}{|\mathcal D_i|} {\sum\limits_{n = 1}^{{|\mathcal D_i|}} \nabla f_i({\boldsymbol{\omega} _i},{\boldsymbol{x}_{in}},{y_{in}})} \; .
\end{split}
\label{Eq:gradientdescent}
\end{equation}

Then,
users uploads the updated local model $\boldsymbol{\omega}_i(t+1)$ to the associated base station for aggregation:
\begin{equation}
\begin{split}
{{\theta _{j}}(t + 1)} = \sum\limits_{i = 1}^N a_{ij} {l_{ij}} \boldsymbol{\omega} _{i}(t + 1) \; .
\end{split}
\label{Eq:globalaggregation}
\end{equation}

To measure the quality of model training performance,
denote $G$ as the averaged test accuracy among all users,
and we have:
\begin{equation}
	G = \frac{1}{N}\sum\nolimits_{i = 1}^N {\sum\nolimits_{j = 1}^M {a_{ij}}{g_{ij}} }\; .
	\label{Eq:testaccuracy}
\end{equation}

As for any user $i$ ,
if it is associated to base station $j$ ($a_{ij}=1$),
the test accuracy on its local test data set $\mathcal D_i$ is as follows:
\begin{equation}
  g_{ij}=\frac{1}{\mathcal |\mathcal D_i|}\sum\nolimits_{n = 1}^{\mathcal |\mathcal D_i|} {\mathbb I(\xi(\boldsymbol{\omega}_i,\boldsymbol{x}_{in})-y_{in}) } \; .
\label{Eq:communicationcost}
\end{equation}
where $\xi(\boldsymbol{\omega}_i,\boldsymbol{x}_{in})$ is predicted label of sample $\boldsymbol{x}_{in}$ based on trained model,
while y is the true label.

\subsection{Communication Model}
At each training iteration, the model uploading from user $i$ to its associated base station $j$ incurs energy cost.
Consider an OFDMA-based system in which base station $j$ provides a total bandwidth $B_j$ to its associated users.
Denote $\beta_{ij}$ as the proportion of bandwidth allocated to user $i$,
then we have:
\begin{equation}
	\begin{split}
		\sum\limits_i {{\beta _{ij}}}  \le 1  \; ,
		&\hspace{0.2cm} (j\in \mathcal M) \; .
	\end{split}
	\label{Eq:beta1}
\end{equation}
Note that user $i$ gets a portion of base station $j$'s bandwidth only if it is associated to station $j$,
then we have:
\begin{equation}
\begin{split}
0 \le {\beta _{ij}} \le a_{ij}  \; ,
  &\hspace{0.2cm} (i\in \mathcal N, j\in \mathcal M) \; .
\end{split}
\label{Eq:beta01}
\end{equation}

Denote $r_{ij}$ as the uploading rate at user $i$, we have:
\begin{equation}
\begin{split}
{r_{ij}} = {\beta_{ij}}{B_j}{\log _2}(1 + \frac{{{h_{ij}}{p_i}}}{{{N_0}}}) \; ,
  &\hspace{0.2cm} (i\in \mathcal N, j\in \mathcal M) \; .
\end{split}
\label{Eq:transmissionrate}
\end{equation}
where $h_{ij}$ is channel gain,
$p_i$ is transmission power and $N_0$ is the Gaussion noise.

Denote $Z_i$ as the  model size of user $i$,
then the energy cost for model uploading can be obtained as:
\begin{equation}
\begin{split}
E_{ij} = \frac{{{Z_i}}}{{{r_{ij}}}}\times {p_{i}} \; ,
  &\hspace{0.2cm} (i\in \mathcal N, j\in \mathcal M) \; .
\end{split}
\label{Eq:energyconsumption}
\end{equation}
Then the system-level energy cost can be obtained as:
\begin{equation}
  E=\frac{1}{N}\sum\nolimits_{i = 1}^N {\sum\nolimits_{j = 1}^M {{a_{ij}}{E_{ij}} }} \; .
\label{Eq:communicationcost}
\end{equation}

\subsection{Problem Formulation}

In this study,
our goal is to design a cost efficient personalized model training scheme.
We employ the trade-off between training performance and communication cost as objective function \cite{Lee20:IWC:group}.
The problem can be formulated as follows:

\begin{center}
	\begin{tabular}{ l l }
		\bf OPT-P&               \\
		max &  $(1-{\mu}){\frac{G}{G_{max}}} -{\mu}{\frac{E} {E_{max}}}$ \\
		s.t
		& Clustered learning constraints: (\ref{Eq:aij1}) -- (\ref{Eq:testaccuracy}) ;\\
		& Communication cost constraints: (\ref{Eq:beta1}) -- (\ref{Eq:communicationcost}) \;.
	\end{tabular}
\end{center}
where $\mu$ is a weighting parameter to adjust the tradeoff between training performance and energy cost,
$G_{max}$ and  $E_{max}$ are normalizing constants to eliminate the impact of different orders of magnitude \cite{Nguyen20:WF-IoT:Mobility-Aware}.
In this formulation,
$a_{ij}$ are binary variables and $\beta _{ij}$ are continuous variables.
The formulated problem falls in the category of
a mixed-integer non-linear program (MINLP),
which is intractable in general\cite{Abichandani13:book:MINLP}.
In the following,
we propose an DRL-based edge association strategy
without disturbing the privacy settings of FL principles.

\section{DRL-based Energy Aware Clustered Federated Learning Scheme}
\label{sec:algorithm}

To comply with privacy protection techniques such as nonlinear parameter encryption,
our objective is to obtain the edge association strategy to cluster users across multiple base stations adaptively based on feedback from base stations, without traversing the model parameters at users.
This can be achieved by transforming problem OPT-P into an Markov decision process (MDP)
with pre-fixed values for channel allocation variables ($\beta_{ij}$).
Then the MDP problem of edge association is defined as follows:
\begin{itemize}
	\item [(1)]
	\textbf{State:} At the beginning of each epoch $k$,
	the system state is defined as
	$\bm S(k) =  \left\{\bm A(k-1),\bm\beta(k),\bm\Delta(k)\right\}$,
	where $\bm A(k-1) = \left\{a_{ij}(k-1),i\in\mathcal{N},j\in\mathcal{M}\right\}$ represents the association results of last epoch,
	$\bm \beta(k) = \left\{\beta_{ij}(k),i\in\mathcal{N},j\in\mathcal{M}\right\}$
	represents the resource allocation solution fed back by base stations,
	$\bm \Delta(k) = \left\{\Delta_i(k),i\in\mathcal{N}\right\}$ is a binary indicator that $\Delta_i(k)=1$ indicates training performance is improved at epoch $k$.
	
	\item [(2)]
	\textbf{Action:} At each epoch $k$,
	the cloud server performs edge association:
	$\bm{A}(k) = \left\{a_{ij}(k),i\in\mathcal{N},j\in\mathcal{M}\right\}$.
\end{itemize}

Due to the test accuracy cannot be calculated in advance,
the state transition probability is difficult to model.
Since the action space is discrete,
we employ double dueling deep Q-learning network (D3QN) based approach to facilitate fast decision making,
which avoids overestimation of Q value by decoupling the action selection and Q value estimation\cite{Wang15:arXiv:duelingDQN},
and further improves training performance in large state and action space.

We employ the objective function of OPT-P as reward function.
Given a weight parameter $\mu$,
the cloud server adopts feedback from each base station on energy consumption ($E$) and model accuracy ($G$) based on the last edge association plan
to guide the clustering process.

As for energy consumption ($E$), it is obtained by solving a bandwidth allocation optimization problem at each edge base station with the aim of minimizing energy consumption among its associated users, which will be described in detail in Sec.~III-A.
Meanwhile,
in order not to interfere with the privacy settings in FL,
we add training performance feedback in terms of average local model accuracy ($G$) to provide an evaluation of the current clustering settings.
Note that in case when users of different conditional distributions are associated to the same base station,
we employ hierarchical transfer learning to improve the training performance,
which will be described in detail in Sec.~III-B.

The current network and the target network are parameterized by $\theta$ and $\theta'$ respectively,
and the target value can be obtains as:

\begin{eqnarray}
&\hspace{-1em}\!{y}\! = \!{(1\!-\!{\mu}){\frac{G}{G_{max}}}\!- \!{\mu}{\frac{E} {E_{max}}}}\! +\! \gamma Q({S},\!\mathop {\arg \max }\limits_{{{A'}} \in {{\bm{A}}}}\! Q({S},\!A',\!\theta ),\!\theta')\!\;.
\label{Eq:proofBound}
\end{eqnarray}

Denote $B$ as mini-batch size and $\nabla_{\theta}$ as gradient,
the update formula for $\theta$ is:
\begin{eqnarray}
	\theta=\theta+\frac{1}{B}\rho[y- Q({S},A,\theta )]\nabla_{\theta}Q({S},A,\theta )\;.
	\label{Eq:D3QNgradient}
\end{eqnarray}


\subsection{Optimal Communication Resource Allocation Strategy}

During each epoch $k$,
given an edge association strategy ($a_{ij}$ are fixed),
each base station $j$ solves an energy consumption minimization problem to obtain bandwidth allocation strategy for its associated users ($\mathcal C_j$).
Then it feeds back the averaged energy consumption ($\frac{1}{|\mathcal C_j|}\sum\limits_{i\in \mathcal C_j} E_{ij}$) as part of reward function to guide the clustering process.
Through simplifying problem OPT-P for a single base station,
each base station $j$ solves the following problem:
\begin{center}
	\begin{tabular}{ l l }
		\bf (P1)&               \\
		min &  $\frac{1}{|\mathcal C_j|}\sum\limits_{i\in \mathcal C_j} E_{ij}$ \\
		s.t
		& Communication cost for users $\in \mathcal C_j$: (\ref{Eq:beta1}) -- (\ref{Eq:energyconsumption}) .
		\label{Eq:allocationproblem1}
	\end{tabular}
\end{center}
In this formulation, $a_{ij}$ are constants and ${\beta_{ij}}$ are optimization variables.
It is easy to prove that problem \textbf{P1} is a convex problem.
We exploit the Karush-Kuhn-Tucker (KKT) conditions for problem \textbf{P1} to obtain the following result.
\begin{theorem} \label{theorem2}
	The optimal bandwidth allocation solution for user $i$ associated to base station $j$ satisfy:
	\begin{equation}
	{\beta _{ij}^{*}} = \frac{{{{\left( {\frac{{{p_i}{Z_{i}}}}{{{|\mathcal C_j|}{B_j}{\log }_2 (1 + \frac{{{h_{ij}}{p_i}}}{{{N_0}}})}}} \right)}^{\frac{1}{2}}}}}{{\sum\limits_{i \in {\mathcal C_j}} {{{\left( {\frac{{{p_i}{Z_{i}}}}{{{|\mathcal C_j|}{B_j}{\log }_2 (1 + \frac{{{h_{ij}}{p_i}}}{{{N_0}}})}}} \right)}^{\frac{1}{2}}}} }} \;.
	\label{Eq:bandwidthsolution}
	\end{equation}

\begin{proof}

The problem \textbf{P1} could be solved by the Lagrange multiplier method.
The Lagrange formula  can be obtained as:
\begin{equation}
L({\beta _{ij}},\lambda ) = \frac{1}{{{|\mathcal C_j|}}}\sum\limits_{i \in {\mathcal C_j}} {{p_i}\frac{{{Z_{i}}}}{{{\beta _{ij}}{B_j}{{\log }_2}(1 + \frac{{{h_{ij}}{p_i}}}{{{N_0}}})}}}  + \lambda (\sum\limits_{i \in {\mathcal C_j}} {{\beta _{ij}}}  - 1)
\label{Eq:lagrangeformula}
\end{equation}
where, $\lambda$ is the Lagrange multiplier of the constraint condition.

In order to obtain the necessary and sufficient conditions of the optimal solution, the KKT conditions  can be obtained as:
\begin{equation}
\frac{{\partial L({\beta _{ij}},\lambda )}}{{\partial {\beta _{ij}}}} = \frac{1}{{{|\mathcal C_j|}}}\frac{{ - {p_i}{Z_{i}}}}{{\beta _{ij}^2{B_j}{{\log }_2}(1 + \frac{{{h_{ij}}{p_i}}}{{{N_0}}})}} + \lambda  = 0
\label{Eq:kkt1}
\end{equation}
\begin{equation}
\lambda (\sum\limits_{i \in {\mathcal C_j}} {{\beta _{ij}}}  - 1) = 0
\label{Eq:kkt2}
\end{equation}

By solving equation (\ref{Eq:kkt1}), we have:
\begin{equation}
\frac{{{p_i}{Z_{i}}}}{{\beta _{ij}^2{B_j}{{\log }_2}(1 + \frac{{{h_{ij}}{p_i}}}{{{N_0}}})}} = \lambda {|\mathcal C_j|}
\label{Eq:solvingequation1}
\end{equation}

Therefore, based on this, we can get the expression of bandwidth allocation and Lagrange Multiplier.Then we have:
\begin{equation}
{\beta _{ij}} = {(\frac{1}{\lambda }\frac{{{p_i}{Z_{i}}}}{{{|\mathcal C_j|}{B_j}{{\log }_2}(1 + \frac{{{h_{ij}}{p_i}}}{{{N_0}}})}})^{\frac{1}{2}}}
\label{Eq:betaintermediatesolution}
\end{equation}

\begin{equation}
\lambda  = \beta _{ij}^2\frac{{{p_i}{Z_{i}}}}{{{|\mathcal C_j|}{B_j}{{\log }_2}(1 + \frac{{{h_{ij}}{p_i}}}{{{N_0}}})}}
\label{Eq:lamdaintermediatesolution}
\end{equation}

Another relation expression according to (\ref{Eq:kkt2}) can be obtained as:

\begin{eqnarray}
\sum\limits_{i \in {\mathcal C_j}} {\beta _{ij}}  = 1
\label{Eq:betasum}
\end{eqnarray}

Furthermore, based on (\ref{Eq:betaintermediatesolution}),(\ref{Eq:lamdaintermediatesolution}),(\ref{Eq:betasum}),  we have:
\begin{equation}
\begin{split}
 {\lambda ^{\frac{1}{2}}} & = \sum\limits_{i \in {\mathcal C_j}} {{{(\frac{{{p_i}{Z_{i}}}}{{{|\mathcal C_j|}{B_j}{{\log }_2}(1 + \frac{{{h_{ij}}{p_i}}}{{{N_0}}})}})}^{\frac{1}{2}}}}\\
&= \frac{1}{{{\beta _{ij}}}}{(\frac{{{p_i}{Z_{i}}}}{{{|\mathcal C_j|}{B_j}{{\log }_2}(1 + \frac{{{h_{ij}}{p_i}}}{{{N_0}}})}})^{\frac{1}{2}}}
\label{Eq:threeexpression}
\end{split}
\end{equation}

Thus, the optimal bandwidth allocation of a single edge base station  under a given edge association strategy can be obtained as:
\begin{equation}
{\beta _{ij}} = \frac{{{{\left( {\frac{{{p_i}{Z_{i}}}}{{{|\mathcal C_j|}{B_j}{\log }_2 (1 + \frac{{{h_{ij}}{p_i}}}{{{N_0}}})}}} \right)}^{\frac{1}{2}}}}}{{\sum\limits_{i \in {\mathcal C_j}} {{{\left( {\frac{{{p_i}{Z_{i}}}}{{{|\mathcal C_j|}{B_j}{\log }_2 (1 + \frac{{{h_{ij}}{p_i}}}{{{N_0}}})}}} \right)}^{\frac{1}{2}}}} }}
\label{Eq:beta}
\end{equation}
\end{proof}
\end{theorem}

\begin{remark}
One observation from Theorem 1 is that more bandwidth should be allocated to users with worse channel.
The reason is that it balances the transmission time required by users with heterogeneous channel states, and thus minimizes the system-level energy consumption under fixed transmission power.
\end{remark}

Then base station $j$ feeds back the averaged energy consumption of all associated users to cloud server,
as a guidance for updating the edge association strategy.

\subsection{Hierarchical Edge Transfer Learning}
During each epoch $k$,
the learning model accuracy is adopted as user feedback ($g_{ij}$), and we use it as part of the reward to guide the clustering process.
In the case when users with different conditional distribution are associated with the same base station due to energy limitations,
we propose to integrating hierarchical transfer learning for such users during training process to further improve its model accuracy.
The details are  as follows:

\noindent\textbf{Identify compromised users.}
After obtaining the edge association solution at each iteration,
we first identify compromised users.
Among base station $j$'s associated users ($\mathcal{C}_j$),
the compromised users are defined as those whose model accuracy are below the averaged accuracy.
As for identified compromised users (denoted as $\mathcal{E}_j$),
we integrate hierarchical transfer learning in their training process.

\noindent\textbf{Hierarchical transfer learning.}
As for any compromised user $i\in \mathcal{E}_j$,
it divides the downloaded personalized mode into base layer (e.g., convolution layer and pooling layer in CNN)
and personalization layer (e.g., fully connected layers) \cite{Arivazhagan19:arXiv:PersonalizationLayers}.
At each training iteration,
user $i$ only uploads the weight-updates for base layer to learn common features (${{\theta _{B_{j}}}(t + 1)} = \sum\limits_{i\in \mathcal C_j} {l_{ij}} {\omega _{B_{j}}}(t + 1)$).
The personalization layer is retained at user to combat statistical heterogeneity,
which is updated locally (${\boldsymbol{\omega} _{P_{i}}}(t + 1) = {{\boldsymbol{\omega}_{P_{i}}}(t)} - \eta \frac{1}{|\mathcal D_i|} {\sum\limits_{n = 1}^{{|\mathcal D_i|}} \nabla f_i({\boldsymbol{\omega} _{P_{i}}},{\boldsymbol{x}_{in}},{y_{in}})}$).
Therefore,
there is no additional overhead.
Note that as for other users $n \in \mathcal{C}_j/\mathcal{E}_j$,
both layers are trained by federated averaging.

\begin{figure*}[htbp]
	\centering
	\captionsetup[figure]{singlelinecheck=off}
	\begin{subfigure}[c]{0.333\textwidth}
		\centering
		\includegraphics[width=1\textwidth]{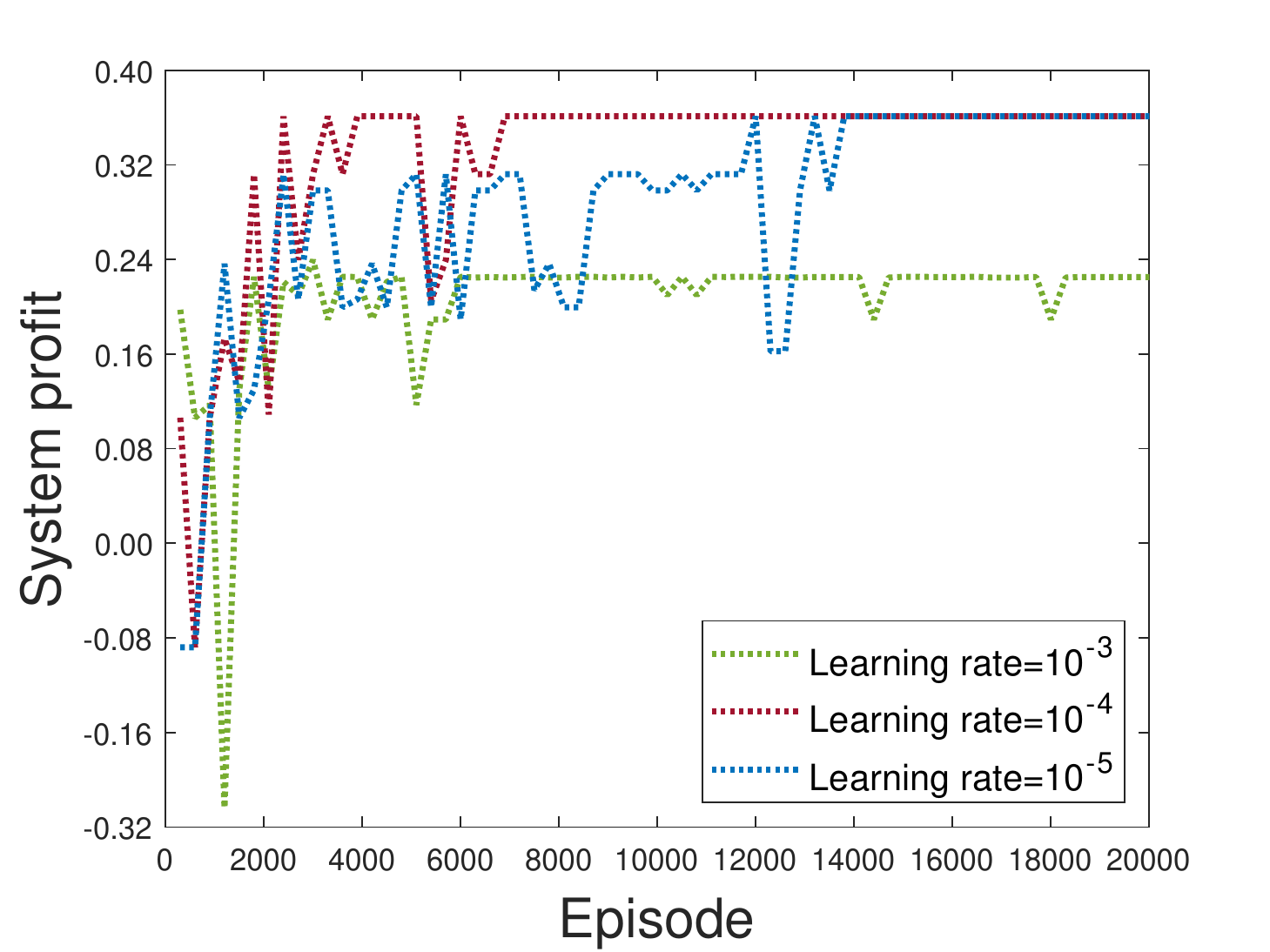}
		\caption{\label{Figure:learning-rate}}
	\end{subfigure}
	\hspace{-0.1in}
	\begin{subfigure}[c]{0.333\textwidth}
		\centering
		\includegraphics[width=1\textwidth]{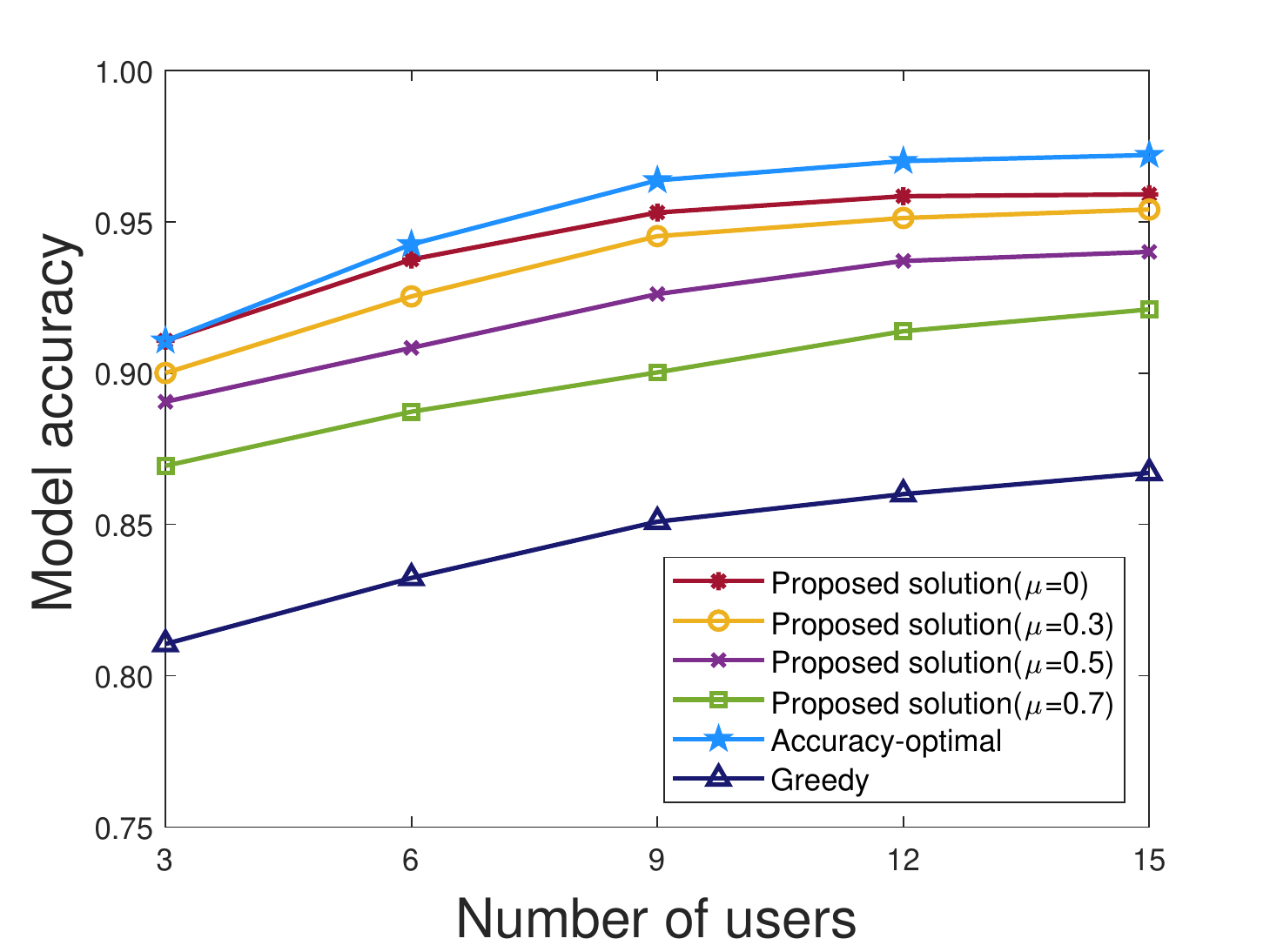}
		\caption{\label{subfigure-accuracy}}
	\end{subfigure}		
	\hspace{-0.1in}	
	\begin{subfigure}[c]{0.333\textwidth}
		
	\centering		
	\includegraphics[width=1\textwidth]{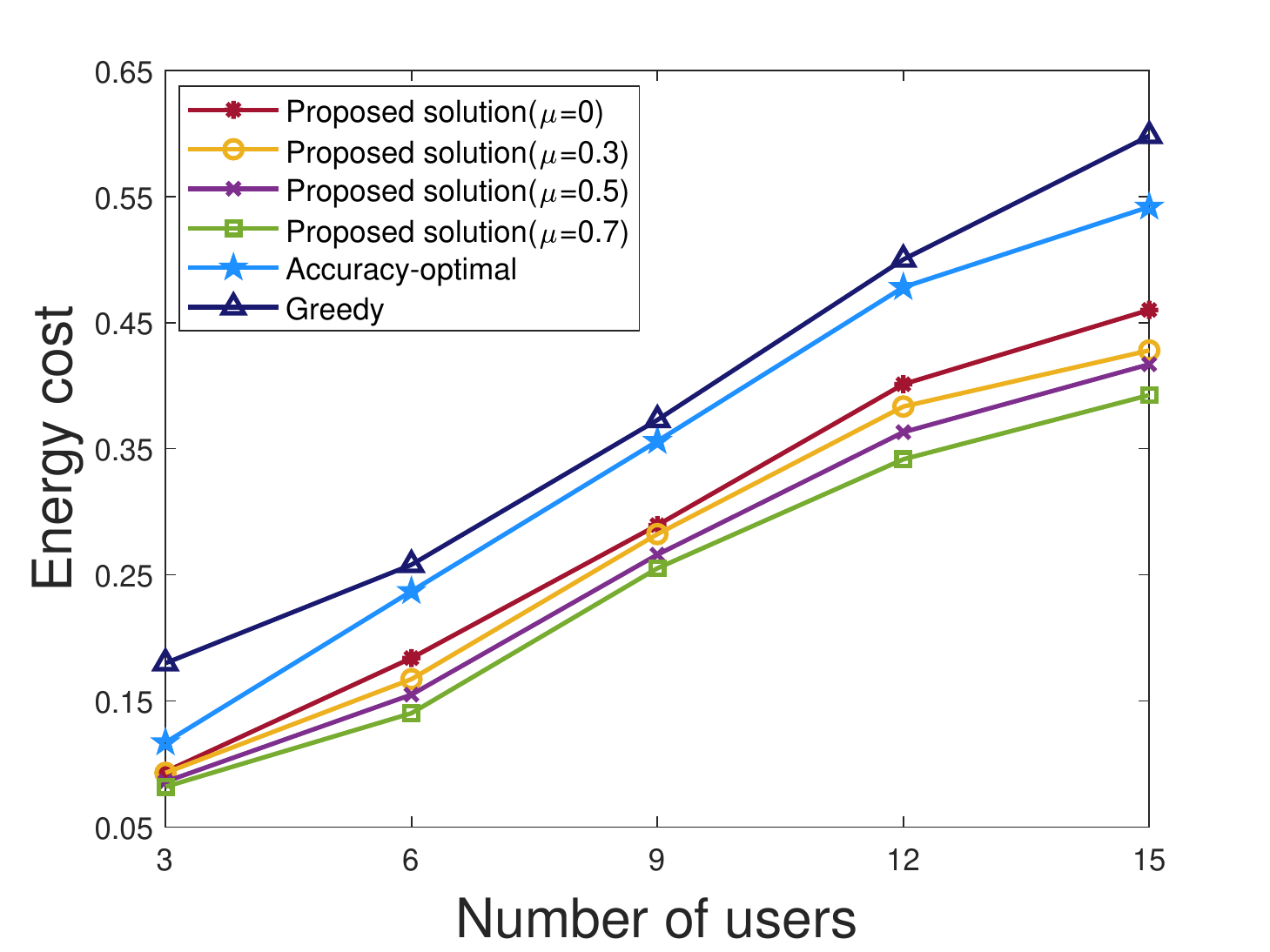}
	\caption{\label{subfigure-energy}}
	\end{subfigure}	
	\caption{(a) Convergence performance under different learning rate. (b) Model accuracy as the number of users increases under different strategies. (c) Energy consumption as the number of users increases under different strategies.}
	\label{Figure:performence}
\end{figure*}

\begin{figure*}[htbp]
	\centering
	\captionsetup[figure]{singlelinecheck=off}
	\begin{subfigure}[c]{0.333\textwidth}
		\centering
		\includegraphics[width=1\textwidth]{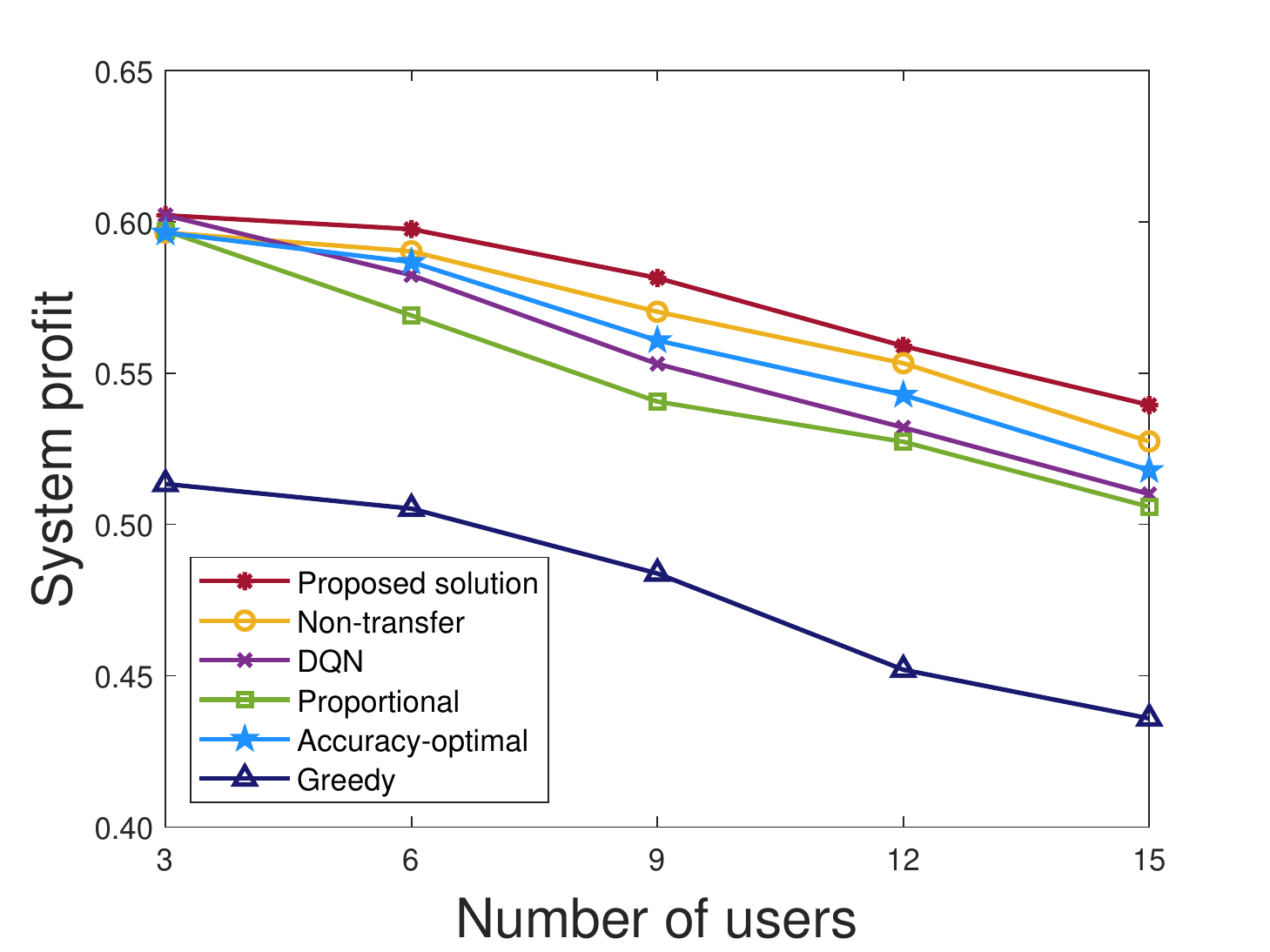}
		\caption{$\mu=0.3$.\label{subfigure_profit03}}
	\end{subfigure}
	\hspace{-0.1in}
	\begin{subfigure}[c]{0.333\textwidth}
		\centering
		\includegraphics[width=1\textwidth]{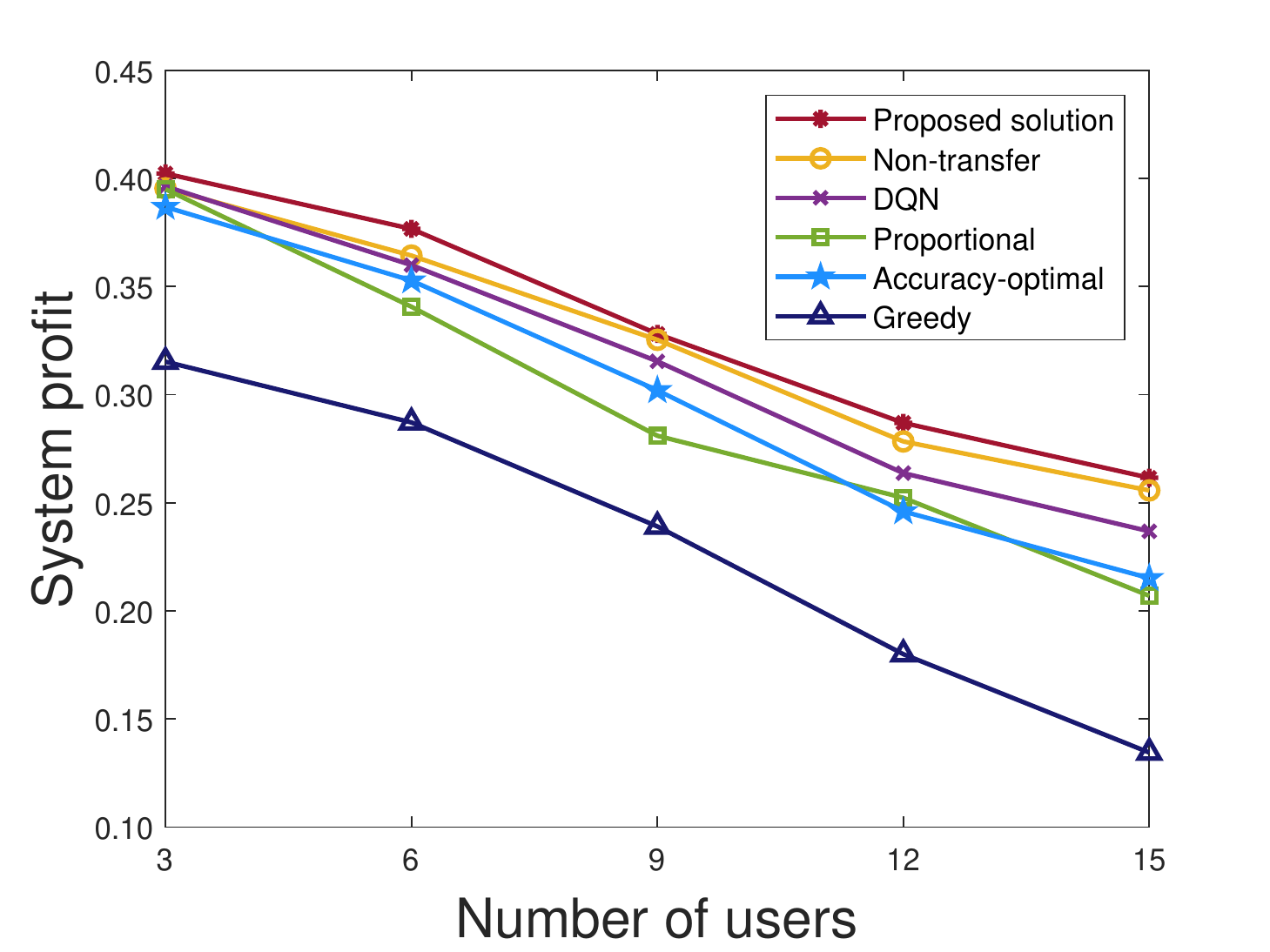}
		\caption{$\mu=0.5$.\label{subfigure_profit05}}
	\end{subfigure}		
	\hspace{-0.1in}	
	\begin{subfigure}[c]{0.333\textwidth}
		
	\centering		
	\includegraphics[width=1\textwidth]{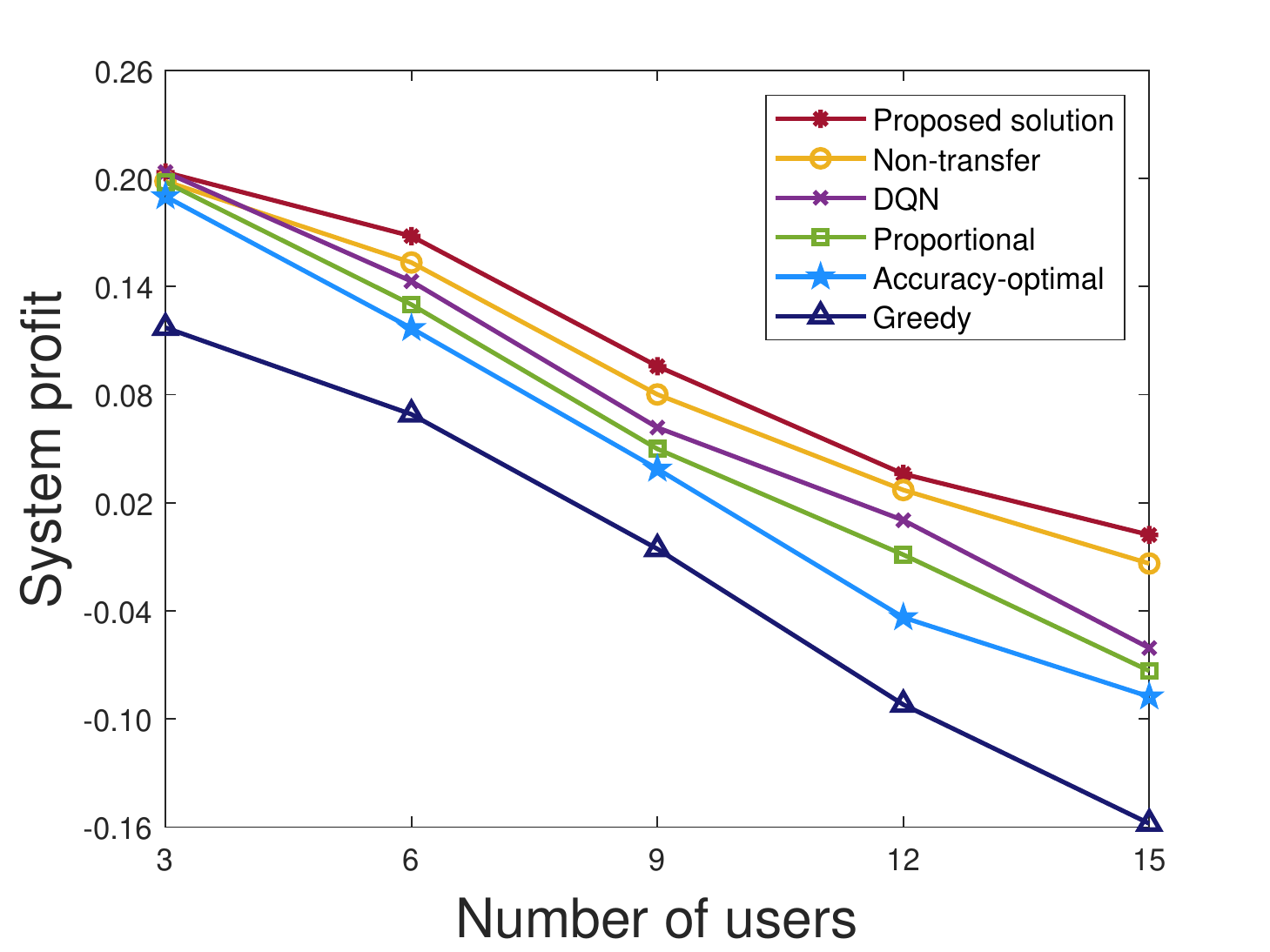}
	\caption{$\mu=0.7$.\label{subfigure_profit07}}
	\end{subfigure}	
	\caption{System profit as the number of users increases under different strategies. }
	\label{Figure:performence}
\end{figure*}

\begin{table*}[!ht]
		{
		\scriptsize
		\begin{center}
			\caption{QUALITATIVE COMPARISON AMONG PROPOSED SOLUTION AND BENCHMARKS}
            \renewcommand\arraystretch{1.4}
			\begin{tabular}{|c|c|c|c|c|c|c|} \hline
				Desideratum                               & Our proposed solution & Non-transfer & DQN  & Proportional  & Accuracy-optimal & Greedy    \\ \hline
				Transfer learning for compromised users   & \Checkmark  & \XSolidBrush  & \Checkmark   & \Checkmark & \XSolidBrush & \Checkmark     \\ \hline
                Avoid Q value overestimation             & \Checkmark  & \Checkmark  & \XSolidBrush & \Checkmark & \XSolidBrush  & \XSolidBrush        \\ \hline
                Optimize bandwidth allocation            & \Checkmark & \Checkmark & \Checkmark & \XSolidBrush & \XSolidBrush & \XSolidBrush       \\ \hline
                No model parameter knowledge required     & \Checkmark & \Checkmark & \Checkmark & \Checkmark& \XSolidBrush& \Checkmark         \\ \hline
                Optimize edge association                & \Checkmark  & \Checkmark & \Checkmark & \Checkmark & \XSolidBrush & \XSolidBrush     \\
				\hline
			\end{tabular}
			\label{table:Simulation}
		\end{center}
	}
\end{table*}

\section{PERFORMANCE EVALUATION}
\label{sec:simulation}

In this section,
we present simulation results to demonstrate the performance of our proposed strategy.
The simulation settings are as follows.
We consider a clustered federated learning system with 2 base stations and 15 users, randomly deployed within a square area.
The uplink bandwidth $B$ is set as $20$ MHz.
The transmission power at each user is set as $23$ dbm,
and Gaussian noise is $-96$ dbm.
The channel condition is quantified as:
$h_{ij} = d_{ij}^{-\alpha}|L_{ij}|$,
the path loss exponent for users $\alpha$ is $4$,
and $L_{ij}\sim\mathcal C\mathcal N(0,8)$ is modeled as complex Gaussian random variables characterizing the Rayleigh fading.
We consider the following learning setup.
The data size of model updates is set as $5$ MB.
We evaluate the training efficiency with handwritten digits classification
task over the MNIST data set with $10$ labels,
which consists of 60000 images.
To simulate diverged conditional distribution among users,
we employ label-swapped non-i.i.d. setting\cite{Briggs20:IJCNN:Clusteredfederated2},
where 500 random selected samples (covering 10 labels) are assigned to each user and two of the labels are swapped.
Users with the same swapped labels are considered to have the same conditional distribution.
As for federated learning,
we employ CNNs with two convolutional layers, two pooling layers, and two fully connected layers.
As for D3QN in our proposed strategy,
it consists of two fully connected hidden layers with 128 neurons.

First,
we investigate the impact of learning rate of D3QN.
The number of users is set as 7 and weight parameter $\mu=0.5$.
Figure~\ref{Figure:learning-rate} shows the trend of
objective value in OPT-P under learning rates of 0.001, 0.0001, and 0.00001, respectively.
As shown in the figure,
a larger learning rate speeds up the convergence,
which leads to a worse objective value.
Moreover, a smaller learning rate gives a better objective value with slower convergence rate.
Therefore,
in the following simulations,
we choose a moderate learning rate of 0.0001.

Next,
we compare the model accuracy ($G/G_{max}$) and energy cost ($E/E_{max}$) under our proposed strategy ($\mu=0.3,0.5,0.7$) with two benchmarks:
``Accuracy-optimal'' and ``Greedy''.
As for ``Accuracy-optimal'',
we assume that the data distribution at users is known in advance for perfect clustering,
and users within a cluster (with the same conditional distribution) are associated to the base station with the closest average distance.
As for ``Greedy'',
each user is greedily associated to base station of the best channel condition,
and the heterogeneous data distributions at users is ignored.
Note that bandwidth allocation in these two benchmarks are reversely proportional to channel condition.

Figure~\ref{subfigure-accuracy} shows the trend of averaged model training performance as the number of users increases from 3 to 15.
It evaluates the system-level accuracy of clustered learning.
As shown in the figure,
without prior knowledge of data distributions,
the accuracy of our clustering solution without knowledge on model parameter is close to ``Accuracy-optimal'',
which has full knowledge of data distribution and thus represents the best model accuracy.
This indicates that our proposed strategy achieves relatively good model accuracy while preserving user privacy.
As the value of $\mu$ increases,
the focus of our strategy design shifts towards energy efficiency,
which decreases the accuracy.
Note that the accuracy of our proposed strategy is always much better than ``Greedy'',
where data distributions is ignored.
It verifies the necessity of considering data distribution heterogeneity among users for edge association.

Figure~\ref{subfigure-energy} shows the trend of averaged energy consumption as the number of users increases from 3 to 15.
It shows the reduction in energy cost under our proposed solution.
As shown in the figure,
the energy consumption of our proposed solution is better than ``Accuracy-optimal'',
which only considers data distribution among users.
This indicates that the geographical distribution of users should not be ignored.
Moreover,
the energy consumption of our proposed strategy is much better than ``Greedy'',
where base station of the best channel condition is associated.
It verifies the importance of load balancing among base stations.
As the value of $\mu$ increases,
the performance gap between our proposed solution and two benchmark strategies increases.

Next,
we show the achievable system-level performance of our proposed strategy under different weight parameters.
The system profit is quantified as the objective value of OPT-P (i.e.,  $(1-{\mu}){\frac{G}{G_{max}}} -{\mu}{\frac{E} {E_{max}}}$).
Maximizing the system profit is equivalent to achieving the best model accuracy with the least energy consumption at users.
Therefore,
a larger value of system profit indicates a  better training strategy.
In addition to the aforementioned three benchmarks,
we demonstrate the necessity of optimizing bandwidth allocation by using ``proportional'' as a benchmark,
which substitutes the optimal allocation solution with proportional allocation.
Moreover,
we demonstrate the performance gain of hierarchical transfer learning by using ``Non-transfer'' as a benchmark.
Further,
we demonstrate the benefit of using D3QN by using ``DQN'' with the same setting as a benchmark.


Figure~\ref{subfigure_profit03}-\ref{subfigure_profit07} shows the trend of achievable system profit as the number of users increases from 3 to 15 when $\mu=0.3,0.5,0.7$, respectively.
As shown in the figures,
under the same accuracy-energy trade-off parameter,
the system profit decreases as the number of users increases.
It indicates that the increase in energy consumption due to limited communication resource is more severe than the benefit in accuracy attributing to more participating users,
which affirms the necessity of incorporating communication cost in strategy design for wireless clustered learning system.
Our proposed strategy achieves the best performance,
while ``Non-transfer'',``DQN'' and ``Proportional'' gives an inferior performance in model accuracy,
edge association and bandwidth allocation respectively.
Note that under same network settings,
the achievable system profit under our proposed D3QN-based approach is always better than DQN-based approach,
and the performance gain increases as the number of users increases.
It shows that D3QN-based approach converges to a better solution as the action space grows.
``Accuracy-optimal'' and ``Greedy'' lacks consideration of geographical distribution and heterogeneity of data distributions,
Due to the inconsistency between user data distribution and geographical distribution,
an edge association solution cannot simultaneously serve the purpose of optimizing energy consumption and optimizing model accuracy.
Therefore,
these two strategies cannot be applied to find the energy-efficient edge association strategy for clustered federated learning.
It verifies the necessity of joint consideration of model accuracy and energy consumption.

\section{Conclusion}
\label{sec:conclusion}
In this paper,
we studied the design of an energy efficient clustered learning strategy over wireless network with multiple base stations.
We formulated a system-level accuracy-energy optimization problem by jointly considering model accuracy,
bandwidth allocation, and energy consumption.
We proposed an iterative solution procedure which involves edge association at cloud server and resource allocation at base stations.
To comply with encryption techniques,
we propose a DRL-based edge association approach to realize clustering without knowledge on model parameters.
Given association decisions,
each base station performs bandwidth allocation using convex optimization.
In the case when users with disagreeing distributions are associated to the same base station due to energy constraints,
we adopt hierarchical transfer learning to further improve the model accuracy.
Simulation results show that the performance of our proposed strategy is competitive in achieving accurate learning at low energy cost.
In future work,
quantization theory can be integrated into our proposed strategy to find a suitable model quantization scheme that further improves the energy efficiency of personalized training under convergence performance guarantees.

\appendices

\section*{Acknowledgment}
This paper is supported by Key Area R\&D Program of Guangdong Province with grant No. 2018B030338001,
Project U2001208 supported by NSFC,
and Beijing Natural Science Foundation under Grant No.L192033.

\ifCLASSOPTIONcaptionsoff
  \newpage
\fi

\end{document}